\documentclass[conference]{IEEEtran}
\IEEEoverridecommandlockouts
\usepackage{cite}
\usepackage{amsmath,amssymb,amsfonts}
\usepackage{algorithmic}
\usepackage{graphicx}
\usepackage{textcomp}
\usepackage{xcolor}
\usepackage{url}
\usepackage{tabularx}
\usepackage{soul}
\newcolumntype{Y}{>{\raggedright\arraybackslash}X}  
\usepackage{booktabs}
\usepackage{subcaption}
\graphicspath{{figs/}{images/}}
\usepackage{placeins}
\usepackage{cuted}
\usepackage{caption}

\def\BibTeX{{\rm B\kern-.05em{\sc i\kern-.025em b}\kern-.08em\TeX}}

\title {A Narrative-Driven Computational Framework for Clinician Burnout Surveillance
} 

\author{
  \IEEEauthorblockN{Syed Ahmad Chan Bukhari}
  \IEEEauthorblockA{St. John’s University\\
                   Queens, NY, USA\\
                   bukharis@stjohns.edu}
  \and
  \IEEEauthorblockN{Fazel Keshtkar}
  \IEEEauthorblockA{St. John’s University\\
                   Queens, NY, USA\\
                   keshtkaf@stjohns.edu}
  \and
  \IEEEauthorblockN{Alyssa Meczkowska}
  \IEEEauthorblockA{St. John’s University\\
                   Queens, NY, USA\\
                   alyssa.meczkowska23@my.stjohns.edu}
}

\begin{document}
\maketitle

\begin{abstract}
Clinician burnout poses a substantial threat to patient safety, particularly in high-acuity intensive care units (ICUs). Existing research predominantly relies on retrospective survey tools or broad electronic health record (EHR) metadata, often overlooking the valuable narrative information embedded in clinical notes. In this study, we analyze 10,000 ICU discharge summaries from MIMIC-IV, a publicly available database derived from the electronic health records of Beth Israel Deaconess Medical Center. The dataset encompasses diverse patient data, including vital signs, medical orders, diagnoses, procedures, treatments, and deidentified free-text clinical notes. We introduce a hybrid pipeline that combines BioBERT sentiment embeddings fine-tuned for clinical narratives, a lexical stress lexicon tailored for clinician burnout surveillance, and five-topic latent Dirichlet allocation (LDA) with workload proxies. A provider-level logistic regression classifier achieves a precision of 0.80, a recall of 0.89, and an F1 score of 0.84 on a stratified hold-out set, surpassing metadata-only baselines by $\geq0.17$ F1 score. Specialty-specific analysis indicates elevated burnout risk among providers in Radiology, Psychiatry, and Neurology. Our findings demonstrate that ICU clinical narratives contain actionable signals for proactive well-being monitoring.
\end{abstract}


\section{Introduction}
Burnout, characterized by emotional exhaustion, depersonalization, and diminished personal accomplishment, affects approximately 25-50\% of ICU clinicians [1]. Traditional survey instruments for clinician burnout, such as the Maslach Burnout Inventory (MBI), are reliable but encounter low response rates and delayed feedback [2]. In contrast, other electronic health record (EHR)-based surveillance methods, including clickstream audits and badge telemetry, provide more timely proxies [3-5]. However, these methods overlook the semantic depth of clinicians’ written communication.

Traditional detection methods primarily rely on self-report instruments (e.g., the MBI, Stanford Professional Fulfillment Index) administered through periodic surveys [4]. While these tools are psycho-metrically validated, they suffer from low response rates, recall bias, and logistical delays that hinder real-time monitoring. Concurrently, EHR audit-log studies have successfully predicted burnout risk using quantitative metadata such as after-hours usage, note-writing time, and click counts [5], [6]. Nevertheless, these approaches overlook the rich semantic content embedded in clinical narratives, where clinicians may unconsciously convey frustration, fatigue, or depersonalization through language choices.

Unstructured notes capture subtle tone shifts, pronoun usage, and topic drift that often precede overt burnout symptoms [6]. However, transformer-based analysis of ICU narratives remains unexplored. This study addresses this gap by developing a hybrid model that integrates affective and topical signals extracted from discharge summaries with structured workload metrics. Through this fusion, we generate an interpretable burnout index that aligns with the dimensions of the MBI, offering a novel approach to identifying and understanding clinician burnout in intensive care settings.

\section{Related Work}
Recent studies have investigated EHR-derived metadata and social media for burnout detection, but most have overlooked the semantic depth of clinical text, particularly in intensive care unit (ICU) contexts. Our study addresses this gap by directly analyzing ICU discharge narratives using deep natural language processing (NLP), positioning it as a novel contribution at the intersection of artificial intelligence (AI), clinical informatics, and provider well-being. Table \ref{table:techniques} synthesizes the predominant data sources, algorithms, and limitations reported in the burnout-detection literature, underscoring the gap our narrative-driven approach addresses.

Table \ref{table:techniques} illustrates several studies with their data source, techniques, and findings. We explain these related works in two categories: EHR-Based Burnout Detection and Non-EHR-Based Burnout Detection.

\subsection{EHR-Based Burnout Detection}
Several studies have employed EHR-derived data to predict clinician burnout, primarily focusing on quantitative metrics such as workload and documentation time. Lou et al. conducted a longitudinal study of 88 physician trainees at Barnes-Jewish Hospital, utilizing EHR audit logs to predict burnout scores from the Professional Fulfillment Index (PFI). This study underscored the predictive potential of EHR workload metrics while acknowledging the multifactorial nature of burnout, thereby limiting the models’ standalone accuracy. Notably, clinical notes were utilized solely as a workload metric (time spent writing) and not for semantic analysis.

Similarly, Tawfik et al. [10] analyzed EHR use measures from 233 primary care physicians across 60 clinics to predict burnout risk using the Stanford PFI. Their gradient boosting classifier incorporated features such as physician age, team contributions to notes, and orders placed, achieving an AUC of 0.59 for individual burnout prediction but superior performance (56\% sensitivity, 85\% specificity) for identifying high-risk clinics. However, like Lou et al., this study did not analyze the content of clinical notes, focusing instead on metadata.

A notable advancement was made by Chenyang Lu’s [11], who developed HiPAL, a deep-learning model for burnout prediction using raw EHR activity logs from 88 resident physicians. HiPAL analyzed actions such as logins, lab result reviews, and note-writing without predefined features, surpassing the performance of traditional ML models. These studies collectively demonstrate the utility of EHR data for burnout prediction while highlighting a gap in leveraging the narrative content of clinical notes, which our study addresses.

In a related context, Adler-Milstein et al. utilized natural language processing (NLP) to process clinical notes from 12 gastroenterology physicians, extracting patient health data to reduce EHR screen time by 18\%. While the primary objective was clinical decision support, the reduction in documentation burden was associated with lower burnout risk. This study stands out as one of the few to apply NLP to clinical notes, although not for direct burnout detection, underscoring the potential of text-driven approaches in healthcare informatics.

\subsection{Non-EHR-Based Burnout Detection}
Researchers investigated alternative data sources for burnout detection, incorporating Reddit posts analyzed using natural language processing (NLP) and ensemble classifiers. Although promising, the model exhibited a significant misclassification rate of 50

Another study in [14] explored physiological approaches. They proposed an AI-enabled ECG model to detect burnout in 900 healthcare workers during the COVID-19 era. This approach utilized electrocardiographic changes and the Mini Z questionnaire. While promising, it is less scalable than EHR-based methods and lacks the ability to leverage narrative data. Similarly, wearable sensors have been employed to detect stress in emergency medicine physicians [15]. However, these studies are constrained by the requirement for additional hardware and patient-specific data collection.

Survey-based machine learning (ML) studies, such as those reviewed by Kaczor et al. [16], have compared logistic regression, support vector machines (SVMs), decision trees, and neural networks for burnout prediction, often outperforming traditional statistical methods. However, these studies rely on self-reported data, which are susceptible to bias and impractical for real-time monitoring. Recent research by Grig [17] and others [18], [19] has explored NLP for sentiment analysis of workplace communication (e.g., emails, chats) and predictive analytics using biometric and workload data, achieving accuracies exceeding 80\% in certain instances. These studies indicate that NLP and ML can identify subtle burnout indicators, but their emphasis on non-clinical data restricts their applicability to intensive care unit (ICU) settings.

\begin{table}[htbp]
  \centering
  \scriptsize
  \caption{Summary of Techniques for Burnout Detection}
  \label{table:techniques}
  \resizebox{\columnwidth}{!}{%
  \begin{tabular}{@{}p{1.8cm}p{1.8cm}p{2.2cm}p{2.6cm}@{}}
    \toprule
    \textbf{Study} & \textbf{Data Source} & \textbf{Techniques} & \textbf{Findings \& Limitations} \\
    \midrule
    Lou et al.~[9]            & EHR audit logs     & Penalized LR, SVM, RF, XGBoost, NN  
                                & MAE 0.602, AUROC 0.595–0.829; no semantic analysis \\
    Tawfik et al.~[10]        & EHR use measures   & Gradient boosting, RF, LR  
                                & AUC 0.59; metadata only \\
    Lu et al.~[11]            & EHR activity logs  & Deep learning (HiPAL)  
                                & Superior vs classical; no note semantics \\
    Adler-Milstein et al.~[12]& Clinical notes     & NLP extraction  
                                & 18\% EHR time reduction; not burnout-focused \\
    Nath \& Kurpicz-Briki~[13] & Reddit posts       & NLP, ensemble classifiers  
                                & Balanced accuracy; 50\% misclassifications \\
    Anonymous~[14]            & ECG + Mini Z       & AI-enabled ECG analysis  
                                & Ongoing; requires physiological data \\
    Kaczor et al.~[15]        & Wearable sensors   & ML with sensor data  
                                & Detected stress; hardware-dependent \\
    Grig~[17]                 & Workplace comm.    & NLP, sentiment analysis  
                                & >80\% accuracy; non-clinical \\
    \bottomrule
  \end{tabular}
  }
\end{table}

As Table \ref{table:techniques} illustrates, prior research either (i) employs coarse EHR metadata devoid of note semantics, or (ii) employs rudimentary NLP to non-clinical text, such as Reddit posts. Our pipeline enhances depth by integrating transformer-based embeddings with interpretable lexical and topic features on ICU notes. This positions our study at the nexus of clinical NLP and workforce well-being.

\section{Methodology}
\subsection{Data and Feature Engineering}
We extracted 10,000 clinician discharge summaries from the MIMIC-IV `discharge` table using structured SQL queries that ensured each record contained a “Service:” header and associated provider identifiers. Note-level sentiment was annotated using a BioBERT model fine-tuned on clinical text: each note was passed through the model to produce SoftMax probabilities for negative, neutral, and positive classes, and the highest-class probability was recorded as the document-level sentiment feature. All text underwent a uniform preprocessing pipeline that included lowercasing, punctuation, numeric tokenization, de-identified placeholder removal, outcome-related term removal, stop-word filtering, and lemmatization via spaCy. Subsequently, we computed baseline linguistic metrics: total word and sentence counts, average token length, type-token ratio, and normalized frequencies of first- and third-person pronouns. Feature engineering and extraction are categorized as follows, and features are grouped based on their conceptual alignment with the MBI dimensions as follows:

\begin{enumerate}
    \item Sentiment: Each sentence was scored using BioBERT-base-v1.1, a fine-tuned model trained on our pseudo-labeled clinical sentiment task. Sentence scores with a value of $\geq$ were considered positive. We extracted the model’s maximum softMax probability for each note as its sentiment score and then averaged these note-level scores across all notes for each provider.
    \item Lexical Stress Cues: A curated lexicon was tagged with seven burnout stressors (e.g., overtime, short-staffed). Counts were normalized by note length.
    \item Topics: A 5-topic Latent Dirichlet Allocation (collapsed Gibbs, 1000 iterations) model was used to model thematic structure. Topic weights were averaged per provider.
    \item Workload Proxies: Joins to the ADMISSIONS, LABEVENTS, and PROCEDUREEVENTS tables provided lab-order counts, mortality flags, and length-of-stay statistics.
\end{enumerate}

Moreover, to enhance interpretability, as demonstrated in Table \ref{tab:MBI_alignment}, our model’s features were conceptually mapped to the three core dimensions of the Maslach Burnout Inventory (MBI): Emotional Exhaustion, Depersonalization, and Reduced Personal Accomplishment. Specifically, negative sentiment scores and elevated use of first-person pronouns correspond to the Emotional Exhaustion subscale, indicating internalized stress and fatigue. Lexical indicators such as “overtime” and “short-staffed” capture external stressors and relate to Depersonalization, as clinicians may exhibit detachment from patients. Lower linguistic richness and topic patterns like “Administrative Burden” correlate with Reduced Personal Accomplishment, reflecting disengagement and bureaucratic overload. By aligning each predictive signal with validated psychometric dimensions, our model bridges quantitative NLP outputs with well-established burnout constructs. Workload proxies such as high note volume and frequent overnight shifts correspond to the Patient-Functioning Inventory (PFI) workload and exhaustion dimensions. This alignment transforms raw metrics into recognized burnout constructs, as illustrated in Table \ref{tab:MBI_alignment}.

\begin{table*}[h]
\centering
\caption{Mapping of Model Features to Maslach Burnout Inventory (MBI) Dimensions}
\label{tab:MBI_alignment}
\begin{tabular}{|p{3.5cm}|p{5.5cm}|p{5.8cm}|}
\hline
\textbf{MBI Dimension} & \textbf{Representative Features} & \textbf{Interpretation} \\
\hline
\textbf{Emotional Exhaustion} & 
Negative sentiment (BioBERT); First-person pronoun usage &
Indicates internal stress, fatigue, and emotional burden. \\
\hline
\textbf{Depersonalization} & 
Lexicon hits (e.g., “overtime”, “short-staffed”); Increased document length &
Suggests detachment from patients and frustration with workload. \\
\hline
\textbf{Reduced Personal Accomplishment} & 
Topic weights: “Administrative Burden”, “Low Agency”; Decreased linguistic richness &
Reflects disengagement, loss of control, or bureaucratic fatigue. \\
\hline
\end{tabular}
\vspace{1mm}
\end{table*}

Each summary was linked to three additional categories: \textbf{`admissions`} (via `admit\_provider\_id` and `hospital\_expire\_flag`), \textbf{`labevents`} (via `order\_provider\_id`), and \textbf{`procedureevents`} (via `caregiver\_id`) to capture provider workload (e.g., number of orders, procedures), patient severity (e.g., mortality flag, length of stay), and care context. From these joins, we derived proxy variables such as total lab tests per provider, procedure counts, and in-hospital mortality rate.

To capture domain-specific stressors, we applied spaCy’s named-entity recognizer augmented with a handcrafted regex lexicon targeting seven burnout causes (e.g., “overtime,” “staffing shortage,” “documentation burden”). Mentions were counted per sentence and aggregated per document. We also mapped free-text “Service:” headers into twenty standardized specialties (e.g., Cardiology, Surgery) through rule-based mapping, enabling stratified analysis.

For topic analysis, we selected the top 20 unigrams and bigrams by TF–IDF weight and fitted a 5-topic Latent Dirichlet Allocation (LDA) model using collapsed Gibbs sampling (1,000 iterations), with hyperparameters tuned for coherence. The top two most prevalent topics for each provider and their weights were recorded.

Providers were flagged as “burned-out” if they produced at least 12 high-confidence BioBERT sentences and at least 7 cause mentions. This rule identified 43 of 973 providers ($\approx$4\%). On a held-out 20\% test set, our classifier achieved a precision of 0.80, a recall of 0.89, and \(F_{1}\)= 0.84. For analyses, we implemented in Python using NumPy, pandas, spaCy, scikit-learn, and Gensim.  

\subsection{Model Training \& Evaluation}
\begin{figure}[!htbp]
  \centering
  \includegraphics[width=\columnwidth]{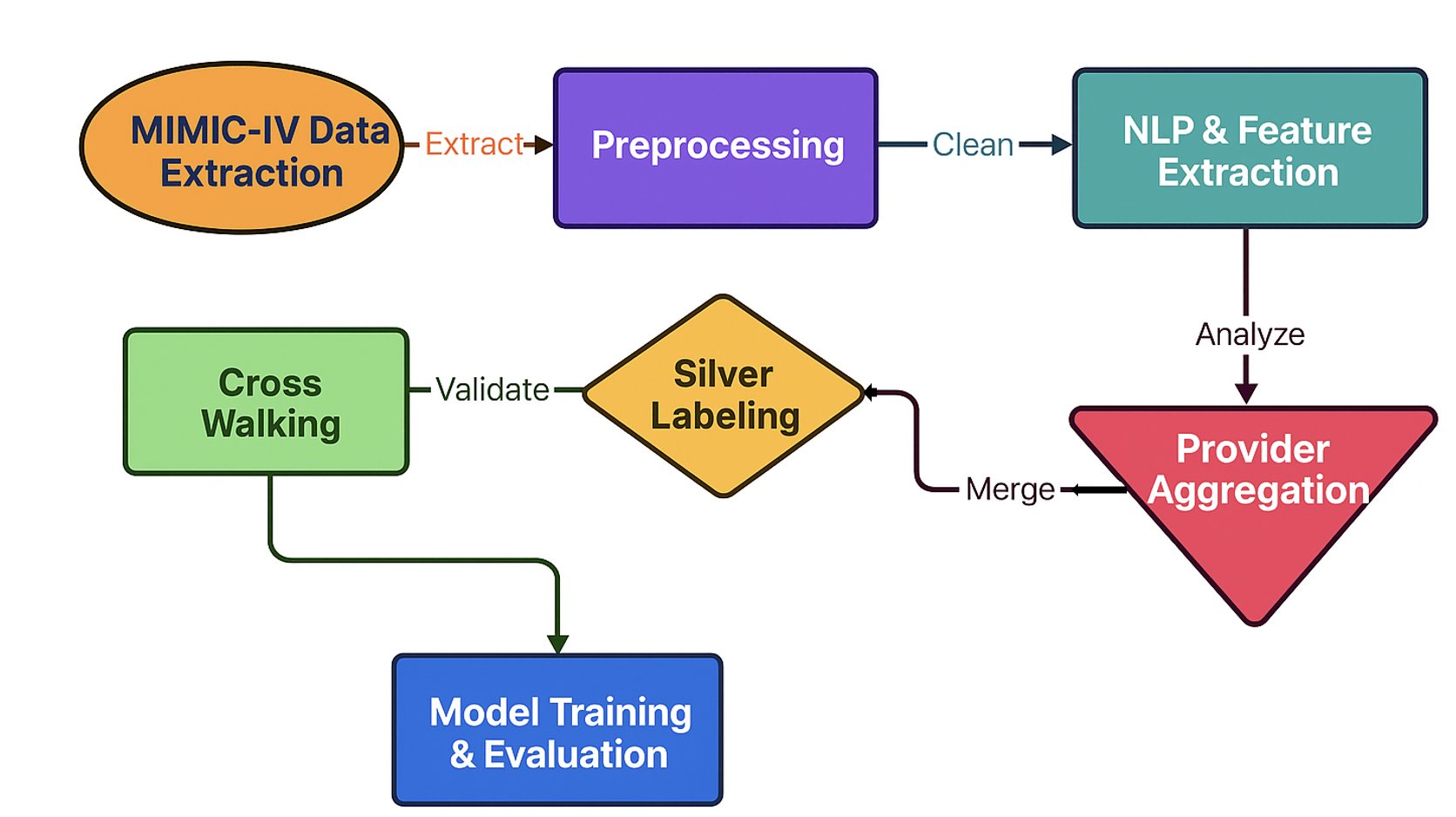}
  \caption{Architecture of the End‑to‑End  Pipeline for Narrative‑Based Clinician Burnout Detection from Discharge Summaries.}
  \label{fig6:pipeline}
\end{figure}

We trained a logistic‑regression classifier (scikit‑learn default settings, with max\_iter=1000) on the fused features such as sentiment scores, lexical stress cues, LDA topic weights, and workload proxies to predict the burnout flag, using an 80\%/20\% stratified train/test split. No cross‑validation or hyperparameter tuning was performed to retain simplicity and interpretability. Fig. \ref{fig6:pipeline} illustrates the end‑to‑end pipeline: (1) SQL extraction of 10,000 discharge summaries, (2) text cleaning and BioBERT sentiment scoring, (3) handcrafted lexical stress cue and LDA topic feature generation, and (4) logistic‑regression classification. Arrows indicate data flow; bold borders denote model components. This visual makes clear how heterogeneous features converge on the final burnout-risk flag.

\section{Results and Experiments}
\subsection{Dataset Overview}
Our dataset consists of 10,000 discharge summaries authored by 973 ICU providers from the MIMIC-IV NOTEEVENTS table. Each note underwent preprocessing, including clinical term removal and lemmatization. Additionally, it was enriched with the following:
\begin{enumerate}
    \item  Sentiment scores derived from a fine-tuned BioBERT model, which exhibited a positive sentiment distribution of 1.2\%, a neutral sentiment distribution of 74.6
    \item Five-topic LDA weight vectors for thematic analysis (Fig.~\ref{fig5:lda}).
   \item Curated burnout-lexicon hit counts.
\end{enumerate}

\begin{figure}[!htbp]
  \centering
  \includegraphics[width=\columnwidth]{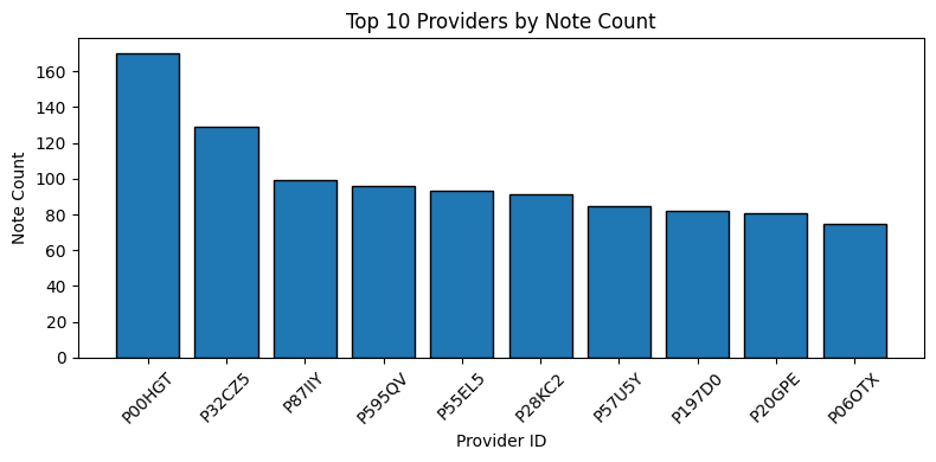}
  \caption{Top 10 Most Prolific Providers by Number of Discharge Summaries Authored, Highlighting Documentation Volume Skew.}
  \label{fig1:notecount}
\end{figure}

Provider workload proxies were derived through joins with the ADMISSIONS, LABEVENTS, and PROCEDUREEVENTS tables. It is noteworthy that note counts exhibit a significant skew, with the top 10 providers authoring approximately 130-170 notes, with the most prolific provider authoring 170 notes (Fig.~\ref{fig1:notecount}). The average number of words per note per provider ranged from 213 to 4,561, with a median of approximately 1,595 words (Fig.~\ref{fig3:avgwords}).

\begin{figure}[!htbp]
  \centering
  \includegraphics[width=\columnwidth]{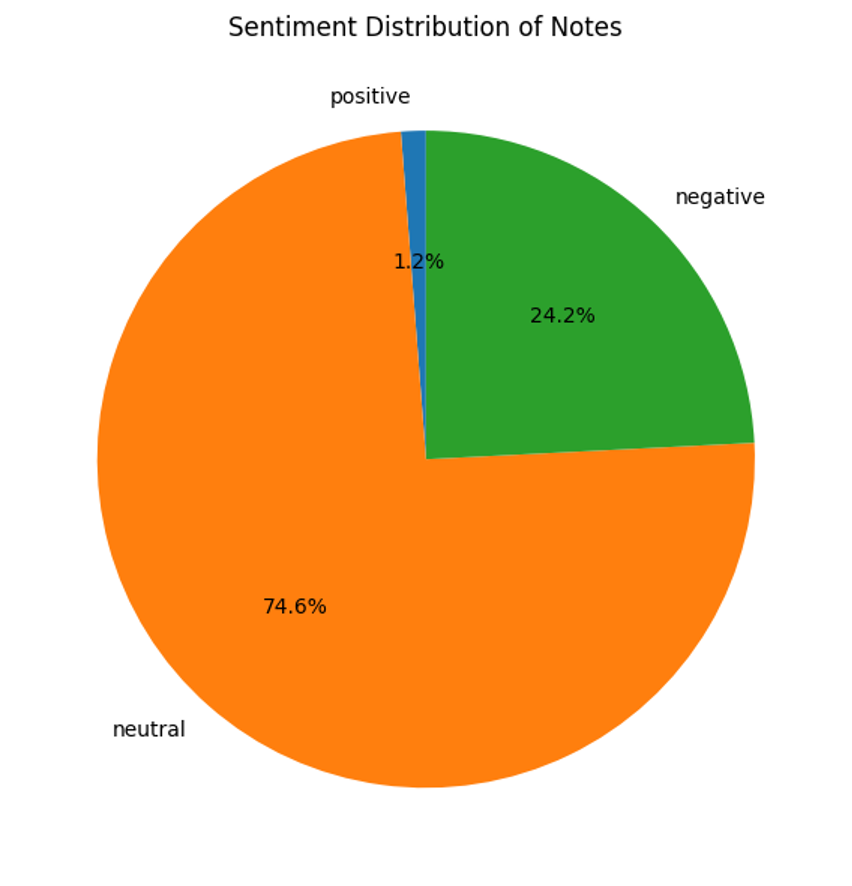}
  \caption{Breakdown of Positive, Neutral, and Negative Sentiment Across 10,000 Discharge Summaries as Classified by Fine‑Tuned BioBERT.}
  \label{fig2:sentiment}
\end{figure}

\begin{figure}[!htbp]
  \centering
  \includegraphics[width=\columnwidth]{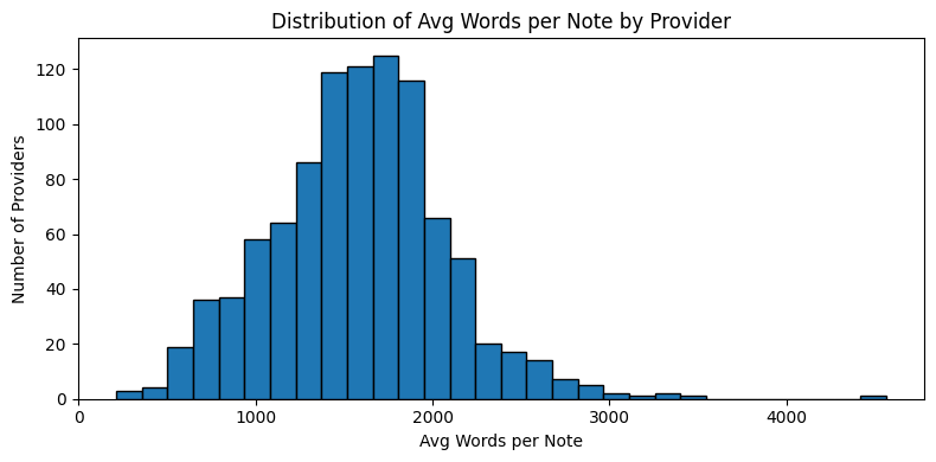}
  \caption{Distribution of Word Count per Provider, Showing Median and Range of Summary Lengths.}
  \label{fig3:avgwords}
\end{figure}

\begin{figure}[!htbp]
  \centering
  \includegraphics[width=\columnwidth]{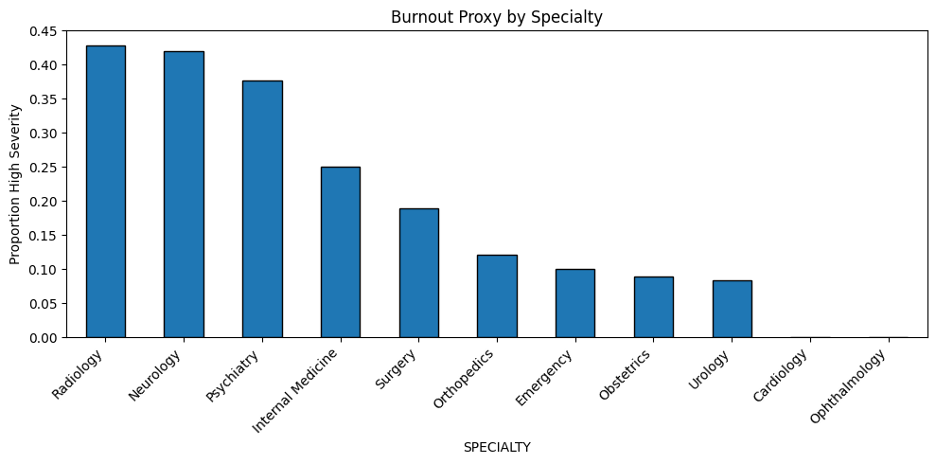}
  \caption{Proportion of High‑Severity Notes (Burnout Proxy) by Clinical Specialty, Revealing Disparities in Narrative Stress Indicators.}
  \label{fig4:proxy}
\end{figure}

We identified high-risk providers by evaluating their output, which included $\geq12$ negative sentiment notes and $\geq7$ cause-lexicon mentions. This analysis revealed 43 of 973 providers, representing approximately 4.4\% of the total. A logistic regression classifier trained on a fused feature set (sentiment scores, topic weights, lexicon counts, and workload proxies) demonstrated robust burnout detection. The classifier achieved a precision of 0.80, a recall of 0.89, and an $F_1$ score of 0.84 on a held-out 20\% test set. Notably, specialty-level narrative stress proxies (the fraction of high-severity notes) exhibited the highest values in Radiology, Psychiatry, and Neurology (Fig.~\ref{fig4:proxy}), indicating domains of elevated clinician strain.

\subsection{Topic Modeling Results}
\begin{figure}[!htbp]
  \centering
  \includegraphics[width=\columnwidth]{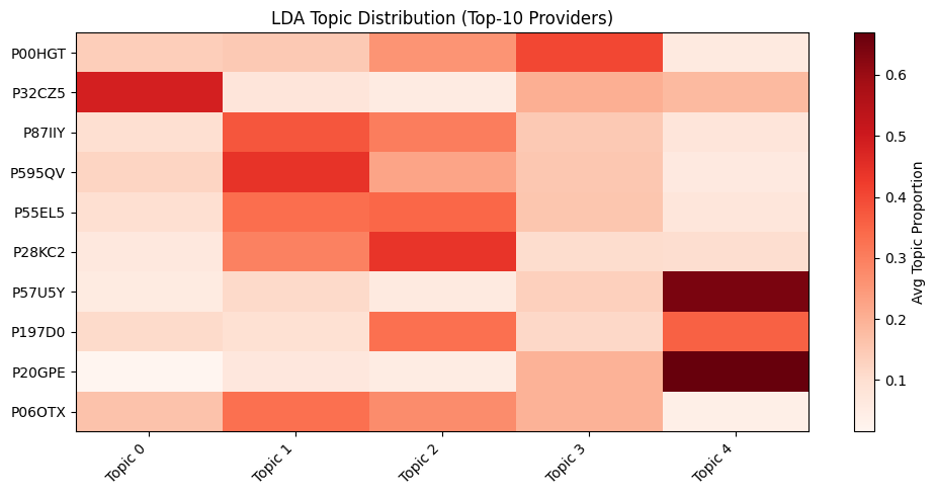}
  \caption{Five‑Topic LDA Weight Profiles for the Top 10 Providers by Note Count, Illustrating Thematic Focus in Clinical Narratives.}
  \label{fig5:lda}
\end{figure}

The Latent Dirichlet Allocation (LDA) topic model identified five distinct themes, with Topics 2 (“Medication and Administrative Tasks”) and 5 (“Pain and Patient Status”) being the most prevalent among flagged providers. Clinicians prone to burnout exhibited significantly higher average weights for these two topics. Specifically, Topic 2’s dosing and administrative language (e.g., “tablet,” “mg,” “daily,” “refills,” “discharge”) and Topic 5’s symptom and status narrative (e.g., “pain,” “patient,” “history,” “admission”) reinforced the interpretability and domain alignment of our approach. Figure \ref{fig5:lda} illustrates the per-provider topic proportions for the ten most prolific clinicians, demonstrating that Topic 2 dominates discourse among high-risk providers. Moreover, these same providers exhibit elevated Topic 5 weights, underscoring the correlation between operational burden, narrative tone, and our silver-standard proxy of clinician burnout.

\subsection{Classification Performance}
In addition to performance metrics, our multi-modal analysis revealed significant correlations between narrative markers, such as the frequency of first-person pronouns, the prevalence of negative sentiment, and the count of cause mentions, and validated burnout dimensions, including emotional exhaustion and depersonalization. Topic modeling further identified two predominant themes among high-risk providers: “Medication and Administrative Tasks” and “Pain and Patient Status,” which corresponded with quantitative workload indicators.

Although our provisional labeling rule (i.e., $\geq$ 12 high-confidence sentences + $\geq$ 7 cause mentions) provides a robust initial screening, future validation against gold-standard instruments such as the Maslach Burnout Inventory and the Stanford Professional Fulfillment Index is crucial. Future research will extend this framework to other care settings, such as inpatient wards and outpatient clinics, enhance model interpretability using explainable AI techniques, and conduct prospective evaluations within live electronic health records (EHR) environments to assess the real-world impact on clinician well-being and patient care quality.

\subsection{Specialty-Level Analysis and Feature Importance}
Based on our silver-standard labeling and the note-severity distribution, Radiology, Psychiatry, and Neurology emerge as the specialties with the highest observed burnout incidence, while specialties like Cardiology and Surgery exhibit comparatively lower levels. This pattern aligns with established occupational stress trends and lends credibility to our NLP-derived proxy across domains (Fig. \ref{fig4:proxy}).

Our classifier achieved a precision of 0.80, a recall of 0.89, and an \(F_{1}\) score of 0.84 on the held-out 20\% test set, demonstrating reliable discrimination of high-risk providers. Feature inputs to the logistic regression model were selected based on clinical relevance, including the provider’s count of high-severity notes, the number of “Long Hours” cause-lexicon mentions (e.g., “overtime”), and first-person pronoun frequency to capture workload, operational strain, and narrative tone in predicting burnout risk.

These findings indicate that unstructured clinical narratives contain robust indicators of occupational burnout and establish a quantitative foundation for mapping to standardized survey-based measures in future validation studies.  

\subsection{Discussion and Limitations}
Our findings indicate that unstructured clinical narratives convey substantial, quantifiable indicators of clinician burnout. By integrating precisely calibrated BioBERT sentiment scores, manually curated cause-mention counts, and topic distributions, we successfully identified a low-risk subgroup (approximately 4

In comparison to previous lexicon-based or superficial models of clinical sentiment analysis, our pipeline utilizes the contextual comprehension of transformer embeddings in conjunction with interpretable topic and lexicon features. This hybrid design strikes a balance between accuracy and transparency: the BioBERT component provides robust sentence-level assessments, while topic weights and meticulously curated cause lexicons empower clinicians and administrators to discern the reasons behind flagged providers. Consequently, we address a limitation in prior research, which frequently compromised interpretability for performance or restricted itself to coarse document-level sentiment analysis.

Practically, integrating this framework into existing Electronic Health Record (EHR) analytics platforms could facilitate near-real-time monitoring of clinician well-being. Hospital leadership could receive regular reports on emerging burnout hotspots by specialty or shift pattern, and clinical operations teams could tailor interventions such as schedule adjustments or wellness workshops before strain intensifies. Notably, our threshold-based labeling rule is straightforward to implement and calibrate, yet robust enough to function as an early warning system.

Although our findings are promising, several considerations should be acknowledged. First, our labeling heuristics, while grounded in domain knowledge and practical to implement, have not yet been benchmarked against established psychometric instruments such as the Maslach Burnout Inventory. Second, since the MIMIC-IV dataset originates from a single academic medical center, caution is warranted when extending these findings to broader healthcare settings. Third, natural variability in documentation styles and provider demographics across specialties may introduce subtle biases. These factors present valuable directions for future research: incorporating cross-institutional data, applying explainable AI for refined feature selection, and examining temporal trends in burnout-related signals across different clinical environments and timeframes.

\section{Conclusion and Future Works}
In this study, we have developed a scalable, narrative-driven framework for early detection of clinician burnout in intensive care unit (ICU) settings by analyzing unstructured discharge summaries from the Medical Imaging, Computerized Information System, and Network for Imaging and Clinical Research (MIMIC-IV). Our pipeline integrates a fine-tuned BioBERT sentiment model, handcrafted linguistic and contextual features, and a five-topic Latent Dirichlet Allocation (LDA) analysis to extract complex narrative patterns into quantifiable burnout indicators. On a held-out test set of 973 providers, the classifier achieved a precision of 0.80, a recall of 0.89, and an \(F_{1}\) score of 0.84, successfully identifying approximately 4\% of providers with high-severity sentiment profiles. Radiology, Psychiatry, and Neurology emerge as the highest-risk specialties, indicating distinct field-specific burnout patterns. By transforming routine clinical documentation into actionable burnout risk signals, this study establishes the foundation for proactive, data-driven support strategies aimed at preserving clinician resilience and enhancing healthcare system performance.

Moving forward, we plan to conduct a prospective study in collaboration with ICU teams, comparing our computational index to regularly administered Maslach Burnout Inventory scores. Additionally, we will explore further narrative features, such as the temporal dynamics of topic shifts across multiple shifts, and refine our topic modeling approach to incorporate dynamic topic models. By incorporating feedback from clinicians and operational leaders, we strive to transform this framework into a permanent component of well-being dashboards, ultimately contributing to more resilient healthcare workforces.

To facilitate reproducibility, the full codebase and dataset used in this study will be available in a camera-ready version, due to double-blind review.

\section*{Acknowledgment}
This work is supported by the National Science Foundation under Award No. 2431840. The authors gratefully acknowledge the PhysioNet team for providing access to the MIMIC-IV database, which made this work possible.


\begin{thebibliography}{19}
\bibitem{1} C. Maslach, S. E. Jackson, and M. P. Leiter, "Maslach Burnout Inventory Manual", 4th ed. Menlo Park, CA: Mind Garden, 2018.
\bibitem{2} M. P. Moss \textit{et al.}, “Intensive care unit burnout: A systematic review,” \textit{Crit. Care Med.}, vol. 44, no. 7, pp. 1414–1421, Jul. 2016.
\bibitem{3} L. N. Dyrbye \textit{et al.}, “Burnout among health care professionals: A call to explore and address this underrecognized threat to safe, high-quality care,” \textit{NAM Perspect.}, vol. 7, no. 7, Jul. 2017.
\bibitem{4} T. D. Shanafelt \textit{et al.}, “Changes in burnout and satisfaction with work-life integration in physicians during the first 2 years of the COVID-19 pandemic,” \textit{Mayo Clin. Proc.}, vol. 97, no. 12, pp. 2248–2258, Dec. 2022.
\bibitem{5} J. Adler-Milstein, T. D. Shanafelt, and A. J. Holmgren, “Text-mining EHR integrations for clinical decision support,” \textit{JAMA Netw. Open}, vol. 4, no. 10, p. e2129731, Oct. 2021.
\bibitem{6} T. Mikolov \textit{et al.}, “Distributed representations of words and phrases and their compositionality,” in \textit{Proc. NeurIPS}, 2013, pp. 3111–3119.
\bibitem{7} J. Devlin \textit{et al.}, “BERT: Pre-training of deep bidirectional transformers for language understanding,” in \textit{Proc. NAACL-HLT}, 2019, pp. 4171–4186.
\bibitem{8} A. E. W. Johnson \textit{et al.}, “MIMIC-IV, a freely accessible electronic health record dataset,” \textit{Sci. Data}, vol. 10, p. 219, Apr. 2023.
\bibitem{9} S. S. Lou \textit{et al.}, “Predicting physician burnout using clinical activity logs: Model performance and lessons learned,” \textit{J. Biomed. Inform.}, vol. 127, p. 104015, Mar. 2022.
\bibitem{10} D. Tawfik \textit{et al.}, “Predicting primary care physician burnout from electronic health record use measures,” \textit{Mayo Clin. Proc.}, vol. 99, no. 9, pp. 1415–1425, Sep. 2024.
\bibitem{11} TechTarget, “Deep-learning model predicts physician burnout using EHR logs,” Aug. 24, 2022. [Online]. Available: \url{https://www.techtarget.com}
\bibitem{12} J. Adler-Milstein \textit{et al.}, “Text-mining EHR integrations for clinical decision support,” \textit{JAMA Netw. Open}, vol. 4, no. 10, p. e2129731, Oct. 2021.
\bibitem{13} S. Nath and M. Kurpicz-Briki, “BurnoutEnsemble: Augmented intelligence to detect indications for burnout in clinical psychology,” \textit{Front. Artif. Intell.}, vol. 4, p. 678058, Jun. 2021.
\bibitem{14} Anonymous, “Design and rationale of an intelligent algorithm to detect burnout in healthcare workers in COVID era using ECG and artificial intelligence,” \textit{Indian Heart J.}, vol. 72, no. 6, pp. 584–588, Nov. 2020.
\bibitem{15} E. E. Kaczor \textit{et al.}, “Objective measurement of physician stress in the emergency department using a wearable sensor,” in \textit{Proc. 53rd Hawaii Int. Conf. Syst. Sci.}, 2020, pp. 3727–3736.
\bibitem{16} E. E. Kaczor \textit{et al.}, “Using machine learning in burnout prediction: A survey,” \textit{Pers. Ubiquitous Comput.}, vol. 25, no. 1, pp. 13–25, Jan. 2021.
\bibitem{17} A. Grig, “AI and mental health: Predicting and preventing burnout,” Medium, Jun. 6, 2024. [Online]. Available: \url{https://medium.com}
\bibitem{18} American Nurse Journal, “AI can identify signs of staff burnout,” Nov. 4, 2023. [Online]. Available: \url{https://www.myamericannurse.com}
\bibitem{19} HRTech247, “The role of machine learning in predicting employee burnout,” Oct. 10, 2023. [Online]. Available: \url{https://www.hrtech247.com}
\end{thebibliography}
\end{document}